\documentclass[english]{amsart}
\usepackage{arxiv}
\usepackage{array}
\usepackage{booktabs}
\usepackage{algorithm2e}
\usepackage{authblk}

\makeatother

\usepackage{babel}
\begin{document}

\title[2323]{Design and implementation of an open source Greek POS Tagger and Entity Recognizer using spaCy}
\author[1]{Eleni Partalidou}
\author[1]{Eleftherios Spyromitros-Xioufis}
\author[1]{Stavros Doropoulos}
\author[2]{Stavros Vologiannidis}
\author[3]{Konstantinos I. Diamantaras}

\affil[1]{DataScouting, 30 Vakchou Street, Thessaloniki, 54629, Greece}

\affil[2]{Dpt of Computer, Informatics and Telecommunications Engineering, International Hellenic University, Terma Magnisias, Serres, 62124, Greece}

\affil[3]{Dpt of Informatics and Electronic Engineering, International Hellenic University, Sindos, Thessaloniki, 57400, Greece}
\maketitle
\pagestyle{plain}

\begin{abstract}
This paper proposes a machine learning approach to part-of-speech
tagging and named entity recognition for Greek, focusing on the extraction
of morphological features and classification of tokens into a small
set of classes for named entities. The architecture model that was
used is introduced. The greek version of the spaCy platform was added
into the source code, a feature that did not exist before our contribution,
and was used for building the models. Additionally, a part of speech
tagger was trained that can detect the morphology of the tokens and
performs higher than the state-of-the-art results when classifying
only the part of speech. For named entity recognition using spaCy,
a model that extends the standard ENAMEX type (organization, location,
person) was built. Certain experiments that were conducted indicate
the need for flexibility in out-of-vocabulary words and there is an
effort for resolving this issue. Finally, the evaluation results are
discussed.
\end{abstract}

\thanks{This work was supported by the Rights, Equality and Citizenship programme
of the European Union (2014-2020).}
\thanks{The content of this publication represents the views of the authors
only and is his/her sole responsibility. The European Commission does
not accept any responsibility for use that may be made of the information
it contains.}

\section{Introduction}

\label{S:1} In the research field of Natural Language Processing
(NLP) there are several tasks that contribute to understanding natural
text. These tasks can manipulate natural language, such as tokenization
process, and consequently can be used in other implementations, in
order to extract syntactic or semantic information. One such task
for syntactic components is Part of Speech Tagging (POS Tagging).
Part of Speech Tagging in corpus linguistics is a process where a
word is assigned with a label of the grammatical term, given the context
it appears in. In many languages, POS Tagging models achieve an accuracy
of 96 to 97 percent \cite{giesbrecht2009ispos}.

Part of Speech Tagging for highly inflective languages, such as Greek
is quite a difficult task. In the Greek Language, words can have different
morphological forms, depending on the part of speech (verbs have up
to ten different forms). For that purpose, there is a need for a tagset
that can support morphological features for improvement of Greek POS
Tagging \cite{prokopidis2011suite}.

Another main task for extracting semantic information is Named Entity
Recognition (NER). Named Entity Recognition is a process where a word
or a set of words reference to a world object. Most Natural Language
Processing models classify named entities that describe people, locations,
organizations, following the ENAMEX type or can be more complex by
detecting numerical types, like percentages (NUMEX) or dates (TIMEX)
\cite{chinchor1999named}.

The greek Part of Speech Tagging and Named Entity Recognition models
presented in this paper were developed using the spaCy library \cite{spacy2}.
SpaCy is an open source, Natural Language Processing library that
supports a variety of tasks, including POS Tagging, Named Entity Recognition,
Dependency Parsing, etc. SpaCy uses sophisticated neural network-based
models for the implementation of Natural Language Processing components
that achieve state-of-the-art results in many of these tasks.

In the following chapters the process for implementing Part of Speech
Tagging and Named Entity Recognition for the Greek Language is explained.
A dataset with extended POS Tags was found and matched to a set of
morphological rules, according to a treebank. The dataset was then
processed, fed to the spaCy model and used for training. Similarly,
for Named Entity Recognition, datasets from different sources were
compared to a custom set of rules for named entities. Finally, different
experiments were conducted for evaluating the accuracy of the models.

\section{SpaCy's deep learning model for POS tagging and Named Entity Recognition}

\label{S:2} SpaCy uses a deep learning formula for implementing NLP
models, summarised as ``embed, encode, attend, predict''. In spaCy's
approach text is inserted in the model in the form of unique numerical
values (ID) for every input that can represent a token of a corpus
or a class of the NLP task (part of speech tag, named entity class).
At the embedding stage, features such as the prefix, the suffix, the
shape and the lowercase form of a word are used for the extraction
of hashed values that reflect word similarities.

At this stage a vocabulary with hashed values and their vectors exist
in the model. For the exploitation of adjacent vectors in the state
of encoding, values pass through the Convolutional Neural Network
(CNN) and get merged with their context. The result of the encoding
process is a matrix of vectors that represents information. Before
the prediction of an ID, the matrix has to be passed through the Attention
Layer of the CNN, using a query vector to summarize the input.

At prediction, a Softmax function is used for the prediction of a
super tag with part of speech and morphology information. Similarly
for named entities, the available class is predicted. After the training
process of the model, the CNN is able to be used for NLP tasks.

In the latest release of spaCy the deep learning models are reported
to be ``10 times smaller, 20\% more accurate and cheaper to run than
the previous generation'' \cite{spacy2}. The models are implemented
using Thinc, spaCy\textquoteright s machine learning library.

\section{Creating a Greek POS Tagger using spaCy}

\label{S:3} The Institute for Language and Speech Processing was
the first to implement a Part of Speech Tagger with morphological
features and has evaluated the experiments in terms of the error rate
of the predicted classes \cite{prokopidis2005theoretical}. These
models can be accessed from web services offered by the Institute
\footnote{http://nlp.ilsp.gr/soaplab2-axis/}. However, the creation
of a compound Greek POS tagger using spaCy, a fast and accurate NLP
python framework is new.

For the creation of a Part of Speech Tagger in the Greek Language
a number of steps was followed. The tags from the ``Makedonia''
dataset, which is described below, were extracted and matched to a
set of morphological rules. The tokens in the dataset were adjusted
to annotation rules that the model will use. Different parameters
in the configuration of spaCy's model were tested while training and
their results are presented in \ref{S:3-4}.

\subsection{Dataset evaluation and selection}

The dataset comes from texts of the Greek newspaper ``Makedonia''.
The articles in the newspaper are categorized in different subjects,
such as sports, health, economy and political news. Data retrieval
was done from the website \footnote{https://inventory.clarin.gr/}
of the clarin project \cite{institute2018modern} and consist of
a set of xml files with information at paragraph, sentence and word
level. It must be underlined that this annotation was performed by
the Institute for Language and Speech Processing and data is licenced
under the CC - BY - NC - SA licence.

Information about the dataset includes the tokens of a set of articles
and their position in a sentence, the lemma and the part of speech
of every token. The various values of POS tags were retrieved and
incorporated into a tag map. The labels and morphology they describe
are explained below.

\subsection{Creation of the Tag Map with reference to Universal Dependencies}

\label{S:3-2} Different labels were found at the dataset and were
matched to a label map, where for each label the part of the speech
and their morphology are analyzed. In more detail, the first two characters
refer to the part of speech and accordingly extend to more information
about it. The label map supports 16 standard part of speech tags:
Adjective, Adposition, Adverb, Coordinating Conjuction, Determiner,
Interjection, Noun, Numeral, Particle, Pronoun, Proper Noun, Punctuation,
Subordinating Conjuction, Symbol, Verb and Other. Each tag describes
morphological features of the word, depending on the part of the speech
to which it refers like the gender, the number, and the case \cite{papageorgiou2000unified}.
It must be mentioned that the extraction of morphological rules and
the matching with the tags was done using the Greek version of the
Universal Dependencies \cite{prokopidis2017universal}.

\subsection{POS Tagger training}

The articles from the newspaper were fed in spaCy library into the
proper format for training. Different parameters were tested, in order
to get the optimal result. The dataset was shuffled, using the same
seed for all the experiments and was split into a train set (70\%),
a test set (20\%) and a validation set (10\%). Information was passed
through the training algorithm in batches with an increasing batch
size from 4 to 32 and a step of 1.001. Additionally, a dropout rate
was configured in every batch, initialized to 0.6 which dropped during
the training process to 0.4. Most of the experiments were trained
using 30 epochs.

The main area of study for the experiments focuses on three important
components. At first, we investigate the difference in results between
part of speech taggers that classify morphological features and taggers
that detect only the part of speech. Moreover, we explore the significance
of pretrained vectors used from a model and their effect on the extraction
of better results. Most importantly, the usage of subwords of tokens
from a tagger as embeddings is issued. For the experiments, precision,
recall and f1 score are used as evaluation metrics.

\subsection{Evaluation and comparison of results}

\label{S:3-4} In the first experiment the model was trained using
pretrained vectors extracted from two different sources, Common Crawl
and Wikipedia and can be found at the official FastText web page \cite{joulin2016bag}.
Both sources were trained on the same algorithm called FastText \cite{grave2018learning},
an extension of Word2Vec that treats tokens as an average sum of sub-words
and finds similarities of words based on their n-grams. The configuration
of the FastText model for Wikipedia vectors is according to \cite{bojanowski2017enriching},
whilst the model for CC vectors is a position-weight CBOW 5 length
n-grams with a window size of 5 tokens and 10 negative words. The
file with the Common Crawl vectors consists of 2.000.000 tokens with
300 dimension, whereas the file with the Wikipedia vectors consists
of 300.000 tokens with 300 dimension.The results can be viewed in
the following table, with the first part describing the Common Crawl
results and the second one the Wikipedia results.

\begin{table}[h]
\begin{tabular}{llll}
\toprule 
\textbf{Classes (Common Crawl)} & \textbf{Precision} & \textbf{Recall} & \textbf{F1 Score}\tabularnewline
\midrule 
POS and morph classes & 96.26 & 96.28 & 96.24\tabularnewline
Only POS classes & 98.75 & 98.75 & 98.75\tabularnewline
\midrule
\textbf{Classes (Wikipedia)} & \textbf{Precision} & \textbf{Recall} & \textbf{F1 Score}\tabularnewline
\midrule
POS and morph classes & 85.90 & 84.27 & 84.46\tabularnewline
Only POS classes & 90.09 & 89.40 & 89.44\tabularnewline
\bottomrule
\end{tabular}

\caption{Results based on Common Crawl pretrained vectors and based on Wikipedia
pretrained vectors}
\end{table}

At the results, POS and morph classes refer to the tag labels explained
in \ref{S:3-2}, whilst only POS classes relate to annotated labels
that describe only the part of speech. It is evident that even though
the CC vectors are noisy, coming from a web source, they lead to better
results than Wikipedia, possibly because they have a larger variety
of tokens.

In the next experiment, the dataset was used for the composition of
embeddings for the part of speech tagger. The dataset was trained
on a FastText model with the same parameters that extracted the Common
Crawl vectors. As a result, 140.000 vectors with 300 dimension were
exported. It must be mentioned that the tagset with the morphological
features was used.

\begin{table}[h]
\centering %
\begin{tabular}{llll}
\hline 
\textbf{Vectors} & \textbf{Precision} & \textbf{Recall} & \textbf{F1 Score}\tabularnewline
\hline 
Common Crawl Vectors & 96.26 & 96.28 & 96.24\tabularnewline
Dataset Vectors & 95.74 & 95.72 & 95.68\tabularnewline
\hline 
\end{tabular}\caption{Usage of pretrained vectors from dataset}
\end{table}

The values of the metrics in this case were almost as good and comparable
to the CC ones. However, the model trained with a larger vocabulary
had higher results. Also, the model with the dataset vectors did not
have the flexibility to classify unknown words.

As a next step, the test set of the dataset was altered by replacing
words with syntactical mistakes to test the tolerance of the model
in OOV words. Suffixes of verbs were altered and vowels were replaced
with others, affecting 20\% of the tokens of the dataset. Using again
the more complex tagset for training, the results can be found in
Table 3.

\begin{table}[h]
\centering %
\begin{tabular}{llll}
\hline 
\textbf{Test Set} & \textbf{Precision} & \textbf{Recall} & \textbf{F1 Score}\tabularnewline
\hline 
Original Test Set & 96.26 & 96.28 & 96.24\tabularnewline
OOV Test Set & 76.40 & 73.77 & 72.22\tabularnewline
\hline 
\end{tabular}\caption{Performance of spaCy model in OOV words}
\end{table}

What can be concluded is that the model did not have a flexibility
in OOV words. Of course, this can also be an advantage, meaning that
the model recognized the mismatch of a wrong word with its class.

One disadvantage that the previous model had is that for unknown words
the model assigned a zero vector, affecting the testing results. In
order to minimize this problem, the unknown words were first passed
through a FastText model to get a vector from their subwords. The
resulting vectors were imported in the vocabulary with the CC vectors
before training. The model was also trained using as a vocabulary
the unknown words and the tokens from the Common Crawl vectors, both
buffered in the same FastText model. Results are listed in Table 4.

\begin{table}[h]
\begin{tabular}{>{\raggedright}p{3.5cm}lll}
\hline 
\textbf{Vectors} & \textbf{Precision} & \textbf{Recall} & \textbf{F1 Score}\tabularnewline
\hline 
Common Crawl Vectors & 96.26 & 96.28 & 96.24\tabularnewline
Common Crawl + FastText(OOV Vectors) & 96.56 & 96.54 & 96.51\tabularnewline
\hline 
Common Crawl Vectors & 96.26 & 96.28 & 96.24\tabularnewline
FastText(Common Crawl + OOV Vectors) & 96.16 & 96.15 & 96.11\tabularnewline
\hline 
\end{tabular}

\caption{Common Crawl pretrained + vectors annotated from out of vocabulary
words and all vectors annotated from FastText (Common Crawl pretrained
and from out of vocabulary words)}
\end{table}

It was noticed that the model performed better when using the vectors
from different FastText models. It was expected that the second experiment
would have performed better, as the tokens were inserted into the
same FastText model and the vectors exported from both sources should
match.

\section{Creating a state of the art Named Entity Recognizer using spaCy}

\label{S:4}In \cite{demiros2000named} the development of an entity
recognizer with named entities that follow a proper set of rules is
described with evaluation metrics that reach 86\% for precision and
81\% for recall. Our implementation follows these rules as well. Also,
a pretrained model is offered from a library called polyglot for recognition
\cite{alrfou2015polyglotner}, which has evaluated NER in Greek with
statistical machine translation.

For the creation of a Named Entity Recognizer in the Greek Language
a number of steps was followed. The entities from the ``Makedonia''
dataset were extracted and annotated, forming a set of keywords that
matched a specific set of rules the entities had to follow. These
keywords were used to reform the dataset and also to find entities
from a larger dataset, like Wikipedia. The spaCy model was trained
using both datasets and their results are compared to a test set.
Additionally, the spaCy model was trained using as a feature the POS
tags of the tokens. All results are presented in \ref{S:4-3}.

\subsection{Dataset evaluation and selection}

In the ``Makedonia'' dataset information about named entities is
organized with the index of the character the named entity starts,
the index of the character the named entity ends and the class of
the named entity. The dataset was parsed and the named entities were
added into the keyword list, with every record representing the token
(or the set of tokens) and its class. Noise was removed from the list
and the records were sorted by the length of the entity. The keyword
list had an average of 72.000 records.

\subsection{Usage of Wikipedia dataset for training}

In order to gain more information about the context of the Greek entities,
a percentage of Greek Wikipedia was used. After applying sentence
and token segmentation on Wikipedia text and using a pretrained model
from polyglot, the keyword list increased. The keyword list had at
this point about 350,000 records and consisted of 4 classes: location
(LOC), organization (ORG), person (PERSON) and facility (FAC). A percentage
of Greek Wikipedia was parsed and used for training in spaCy. The
results from the training are presented in \ref{S:4-3}.

\subsection{Evaluation and comparison of results}

\label{S:4-3} Both datasets were fed into the library in proper format
for training. In training process, the entity recognizer had the same
configuration with the POS tagger, using the same percentages for
train, validation and test sets. It must be noted that all the models
used the Common Crawl pretrained vectors for a vocabulary. The results
are compared using the macro F1 score.

At first the datasets from both sources (Makedonia, Wikipedia) were
used for training with 10 iterations and testing from the model. The
results can be viewed in the following table:

\begin{table}[h]
\begin{tabular}{>{\raggedright}p{3.5cm}l}
\hline 
\textbf{Corpus} & \textbf{Average F1 Score}\tabularnewline
\hline 
Makedonia Corpus & 91.18\tabularnewline
Wikipedia Corpus & 80.75\tabularnewline
\hline 
\end{tabular}\caption{Comparison of Macro Average F1 score with different train sets}
\end{table}

It seemed that the average F1 score was higher for the Makedonia corpus,
as it was the basis of the configuration for the keyword list. In
order to have an objective evaluation, the results of each corpus
per entity class were observed.

\begin{table}[h]
\begin{tabular}{>{\raggedright}p{2.5cm}ll}
\hline 
 & \textbf{Average F1 Score} & \tabularnewline
\hline 
\textbf{Class} & \textbf{Makedonia} & \textbf{Wikipedia}\tabularnewline
\hline 
Non Entity & 0.99 & 0.99\tabularnewline
\hline 
FAC & 0.84 & 0.39\tabularnewline
LOC & 0.94 & 0.88\tabularnewline
\hline 
ORG & 0.93 & 0.88\tabularnewline
PERSON & 0.91 & 0.89\tabularnewline
\end{tabular}\caption{Results of the different train sets per class}
\end{table}

Both sources had good results in non entity tokens, which affected
the F1 score. Moreover, the model did not perform well for facilities,
as polyglot's Greek recognizer does not support that class and FAC
entities cover a small amount of the list.

In the second experiment, the datasets were compared to a common test
set that followed the desired set of rules.

\begin{table}[h]
\begin{tabular}{>{\raggedright}p{3.5cm}l}
\hline 
\textbf{Corpus} & \textbf{Average F1 Score}\tabularnewline
\hline 
Makedonia Corpus & 73.27\tabularnewline
Wikipedia Corpus & 46.47\tabularnewline
\hline 
\end{tabular}\caption{Comparison of results with common test set}
\end{table}

Again, the Makedonia corpus performed better, because of the proper
annotation on the keyword list.

In an experiment worth mentioning the correlation of the part of speech
with the performance of the recognizer was explored. In this experiment,
both pipelines (part of speech, entity recognition) were used for
training with 30 iterations and the model was trained twice: with
and without the usage of the part of speech information for recognition.

\begin{table}[h]
\begin{tabular}{>{\raggedright}p{3.5cm}ll}
\hline 
\textbf{POS feature} & \textbf{Tag F1 Score} & \textbf{Entity F1 Score}\tabularnewline
\hline 
No usage of POS feature & 96.51 & 93.15\tabularnewline
Usage of POS feature & 96.45 & 93.14\tabularnewline
\hline 
\end{tabular}\caption{Results of Makedonia training with discrimination the POS tag as feature
of the model}
\end{table}

It is evident that the recognizer did not gain knowledge from the
part of speech tags of the tokens.

\section{Conclusions}

\label{S:5} Natural Language Processing meets numerous problems in
its applications, especially in uncommon languages such as Greek.
This paper proposes a machine learning approach to part-of-speech
tagging and named entity recognition for Greek, a highly inflected
language using spaCy, a very robust and popular framework. Although
significant work has been done, there are several more things that
can be accomplished. The need of more datasets for the Greek language
is evident, but the results are quite satisfying, comparable to other
languages.
\bibliographystyle{elsarticle-num}
\bibliography{spacy}

\begin{thebibliography}{10}
\expandafter\ifx\csname url\endcsname\relax
  \def\url#1{\texttt{#1}}\fi
\expandafter\ifx\csname urlprefix\endcsname\relax\def\urlprefix{URL }\fi
\expandafter\ifx\csname href\endcsname\relax
  \def\href#1#2{#2} \def\path#1{#1}\fi

\bibitem{giesbrecht2009ispos}
E.~Giesbrecht, S.~Evert, Is {Part}-of-{Speech} {Tagging} a {Solved} {Task}?
  {An} {Evaluation} of {POS} {Taggers} for the {German} {Web} as {Corpus}
  (2009) 9.

\bibitem{prokopidis2011suite}
P.~Prokopidis, B.~Georgantopoulos, H.~Papageorgiou, A {SUITE} {OF} {NATURAL}
  {LANGUAGE} {PROCESSING} {TOOLS} {FOR} {GREEK} (2011) 9.

\bibitem{chinchor1999named}
N.~Chinchor, E.~Brown, L.~Ferro, P.~Robinson, 1999 {Named} {Entity}
  {Recognition} {Task} definition (1999).

\bibitem{spacy2}
M.~Honnibal, I.~Montani, spacy 2: Natural language understanding with bloom
  embeddings, convolutional neural networks and incremental parsing, To appear
  (2017).

\bibitem{prokopidis2005theoretical}
P.~Prokopidis, E.~Desypri, M.~Koutsombogera, H.~Papageorgiou, S.~Piperidis,
  {Theoretical and Practical Issues in the Construction of a Greek Dependency
  Treebank}, in: M.~Civit, S.~Kubler, M.~A. Marti (Eds.), {Proceedings of The
  Fourth Workshop on Treebanks and Linguistic Theories (TLT 2005)}, Universitat
  de Barcelona, Barcelona, Spain, 2005, pp. 149--160.

\bibitem{institute2018modern}
I.~of~Language, S.~Processing, Modern greek texts corpus - "makedonia"
  newspaper annotated by the grne-tagger (2018).

\bibitem{papageorgiou2000unified}
H.~Papageorgiou, P.~Prokopidis, V.~Giouli, S.~Piperidis, A {Unified} {POS}
  {Tagging} {Architecture} and its {Application} to {Greek} (2000) 8.

\bibitem{prokopidis2017universal}
P.~Prokopidis, H.~Papageorgiou, Universal dependencies for greek, in:
  Proceedings of the NoDaLiDa 2017 Workshop on Universal Dependencies (UDW
  2017), Association for Computational Linguistics, Gothenburg, Sweden, 2017,
  pp. 102--106.

\bibitem{joulin2016bag}
A.~Joulin, E.~Grave, P.~Bojanowski, T.~Mikolov, Bag of tricks for efficient
  text classification, arXiv preprint arXiv:1607.01759 (2016).

\bibitem{grave2018learning}
E.~Grave, P.~Bojanowski, P.~Gupta, A.~Joulin, T.~Mikolov,
  \href{http://arxiv.org/abs/1802.06893}{Learning {Word} {Vectors} for 157
  {Languages}}, arXiv:1802.06893 [cs]ArXiv: 1802.06893 (Feb. 2018).
\newline\urlprefix\url{http://arxiv.org/abs/1802.06893}

\bibitem{bojanowski2017enriching}
P.~Bojanowski, E.~Grave, A.~Joulin, T.~Mikolov,
  \href{https://transacl.org/ojs/index.php/tacl/article/view/999}{Enriching
  {Word} {Vectors} with {Subword} {Information}}, Transactions of the
  Association for Computational Linguistics 5~(0) (2017) 135--146.
\newline\urlprefix\url{https://transacl.org/ojs/index.php/tacl/article/view/999}

\bibitem{demiros2000named}
I.~Demiros, S.~Boutsis, V.~Giouli, M.~Liakata, H.~Papageorgiou, S.~Piperidis,
  Named {Entity} {Recognition} in {Greek} {Texts} (2000) 6.

\bibitem{alrfou2015polyglotner}
R.~Al-Rfou, V.~Kulkarni, B.~Perozzi, S.~Skiena, {Polyglot-NER}: Massive
  multilingual named entity recognition, {Proceedings of the 2015 {SIAM}
  International Conference on Data Mining, Vancouver, British Columbia, Canada,
  April 30 - May 2, 2015} (April 2015).

\end{thebibliography}

\end{document}